\documentclass[sigconf]{acmart}

\AtBeginDocument{%
  \providecommand\BibTeX{{%
    \normalfont B\kern-0.5em{\scshape i\kern-0.25em b}\kern-0.8em\TeX}}}

\setcopyright{acmcopyright}
\copyrightyear{2021}
\acmYear{2021}
\acmDOI{10.1145/1122445.1122456}

\acmConference[Woodstock '21]{Woodstock '21: ACM Symposium on Neural
  Gaze Detection}{June 03--05, 2021}{Woodstock, NY}
\acmBooktitle{Woodstock '21: ACM Symposium on Neural Gaze Detection,
  June 03--05, 2021, Woodstock, NY}
\acmPrice{15.00}
\acmISBN{978-1-4503-XXXX-X/18/06}

\usepackage{algorithmic}
\usepackage[ruled]{algorithm2e}
\usepackage{graphicx}
\usepackage{graphics}
\usepackage{textcomp}
\usepackage{subfigure}
\usepackage{tabularx}

\usepackage{amsthm,amsmath,amssymb}
\usepackage{mathrsfs}
\usepackage{multicol}
\usepackage{multirow}
\usepackage{bm}
\usepackage{latexsym}
\usepackage{booktabs}
\usepackage{threeparttable}
\usepackage{epsf}
\usepackage{color,colortbl}
\usepackage{ragged2e}
\usepackage{enumitem}
\usepackage{hyperref}
\usepackage{color}

\begin{document}

\title{A Unified Framework for Campaign Performance Forecasting in Online Display Advertising}


\author{Jun Chen, Cheng Chen, Huayue Zhang, Qing Tan}
\email{ {anzhi.cj,chencheng.cc,huayue.zhy,qing.tan}@alibaba-inc.com}
\affiliation{%
  \institution{Alibaba Group, Beijing, China}
  \streetaddress{P.O. Box 1212}
  \city{Beijing}
  \state{Beijing}
  \country{China}
  \postcode{43017-6221}
}








\renewcommand{\shortauthors}{Anonymous Author}

\begin{abstract}
Online advertising has become one of the most promising businesses and will keep fast growth in the future years. Advertisers usually enjoy the flexibility to choose criteria like target audience, geographic area and bid price when planning an campaign for online display advertising, while they lack forecast information on campaign performance to optimize delivery strategies in advance, resulting in a waste of labour and budget for feedback adjustments. In this paper, we aim to forecast key performance indicators for new campaigns given any certain criteria. Interpretable and accurate results could enable advertisers to manage and optimize their campaign criteria. There are several challenges for this very task. First, platforms usually offer advertisers various criteria when they plan an advertising campaign, it is difficult to estimate campaign performance unifiedly because of the great difference among bidding types. Furthermore, complex strategies applied in bidding system bring great fluctuation on campaign performance, making estimation accuracy an extremely tough problem. 

To address above challenges, we propose a novel Campaign Performance Forecasting framework, which firstly reproduces campaign performance on historical logs under various bidding types with a unified replay algorithm, in which essential auction processes like match and rank are replayed, ensuring the interpretability on forecast results. Then, we innovatively introduce a multi-task learning method to calibrate the deviation of estimation brought by hard-to-reproduce bidding strategies in replay. The method captures mixture calibration patterns among related forecast indicators to map the estimated results to the true ones, improving both accuracy and efficiency significantly. To the best of our knowledge, this is the first systematic work for Campaign Performance Forecasting, and it could be well generalized to other platforms. Experiment results on a dataset from Taobao.com demonstrate that the proposed framework significantly outperforms other baselines by a large margin, and an online A/B test verifies its effectiveness in the real world. We deploy our framework as daily service on Taobao.com for advertising optimization, which responds to advertisers' requests in seconds, meeting both interpretability and accuracy requirements.
\end{abstract}





\maketitle

\section{Introduction}

Online advertising has become one of the most important business. Worldwide digital advertising expenditure is estimated to be 385 billion U.S. dollars in 2020 and it would further grow to 517 billion by the end of 2023 \cite{advertisingspending}. Real-Time Bidding (RTB)\cite{wang2015real} is the most popular paradigm in online advertising, in which advertisers bid for the ad opportunity at the impression level, with the flexibility to choose when, where, to whom their ads should be shown. The combination of these criteria made by advertisers is also known as an advertising campaign.

Advertisers plan their campaigns on an advertising platform for business promotion, and they usually observe campaign performance to make adjustments after the campaign has delivered for a period of time, like one day, while it is a great waste of labour and budget, especially when an advertiser is a novice. Advertising platform aims to streamline and centralize the processes of planning, executing, optimizing on campaigns. Apparently, a well-performed forecast service on campaign performance provides insight for advertisers to make decisions in advance, minimize the trial cost, thus strengthens advertiser stickiness and attracts more budget for platforms. For further discussion, we define Campaign Performance Forecasting, which estimates key performance indicators (e.g. impression, cost, click etc.) for a new campaign after advertisers specify their criteria.  When advertisers adjust their campaign criteria continuously, forecast results are fed back to advertisers in time for campaign optimization. Therefore, the forecast results should be in accord with common sense on any criteria dimension. For example, the results shouldn't decrease after adding an extra budget, bidprice or targeting crowd etc. We call it as $interpretability$ here, which makes the task differ from a regular end-to-end regression one.


In consideration of the business value Campaign Performance Forecasting brings, most global advertising platforms such as Google, Facebook and Taobao offer forecast tools for advertisers, but no research paper or implementation detail has been released. Public patents\cite{cui2013campaign,kalish2016method,wang2012forecasting,jiang2015predicting} are also too obsolete for employment with the development of online advertising. For example, Facebook \cite{jiang2015predicting} proposed an estimation idea for Cost-Per-Mille(CPM)\cite{asdemir2012pricing} bidding, which reproduces match and rank phase of a bidding system on historical logs. We call the idea as $replay$ in this paper. Specifically, given campaign criteria, it first retrieves complete auctions from historical logs(match), and then determines the number and price charged of previous auctions in which the campaign would win(rank). For $replay$, the same procedure as online bidding system ensures the $interpretability$ on forecast results naturally. However, the method above only works on Cost-Per-Mille(CPM) bidding for forecasting winning impressions and budget spend, while advertisers usually concern more about clicks or conversions in Cost-Per-Click(CPC)\cite{asdemir2012pricing} bidding and Cost-Per-Action(CPA)\cite{hu2016incentive} bidding. As for academic works, there is few work on this problem to the best of out knowledge. The most related ones\cite{shi2018audience,sinha2019forecasting} focus on forecasting granular audience size for DSPs\cite{yuan2013real}. Obviously, they neglect key performance indicators forecast for a campaign.

\begin{figure}[!htp]
\centering
\includegraphics[width=0.475\textwidth]{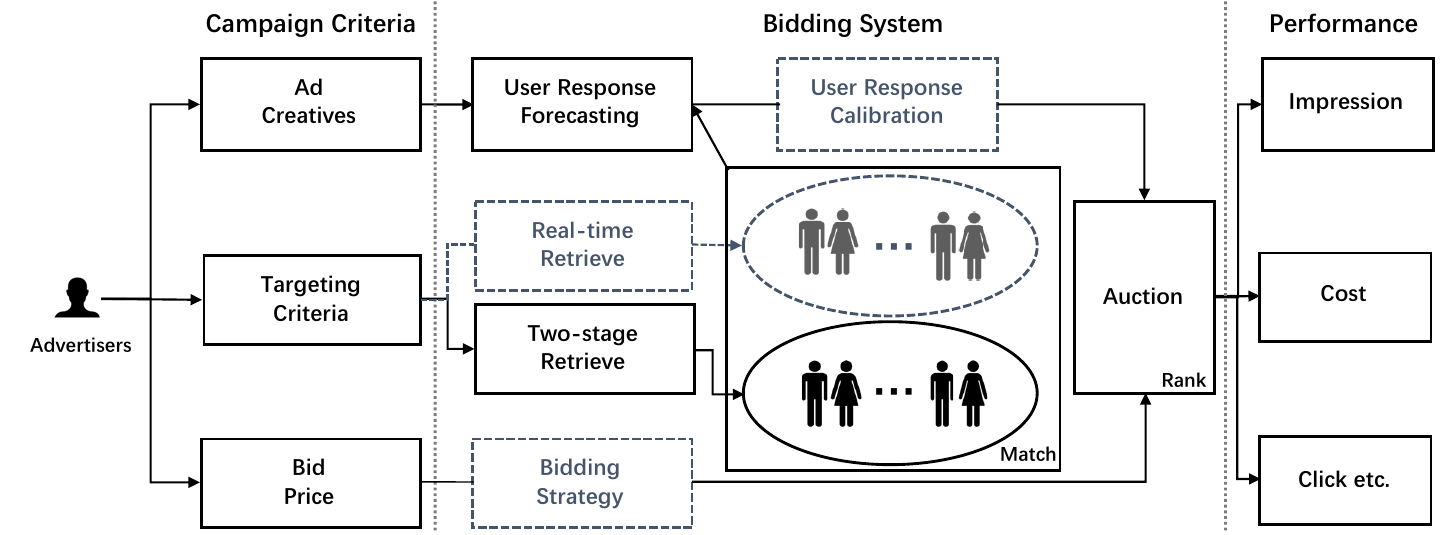}
\caption{A block diagram of common bidding process for an advertising campaign on Taobao.com. Blocks in gray are strategies widely applied on most global advertising platforms.}
\label{auction_process} 
\end{figure}

Nowadays, most advertising platforms provide advertisers various bidding types to achieve personalized advertising, e.g. manual bidding like Cost-Per-Mille(CPM), Cost-Per-Click(CPC) and Cost-Per-Action(CPA), automatic bidding like Budget-Constrained-Bidding(BCB)\cite{wu2018budget} and Multivariable-Constrained-Bidding(MCB) \cite{yang2019bid}, constituting a complex bidding environment. Compared to traditional manual bidding, automatic bidding gets increasingly popular among advertisers, in the bidding process of which user response information plays an important role. The complex bidding environment raises the need of a unified campaign performance forecasting service, which enables advertisers to optimize strategies in time under various bidding types. However, no prior work studies performance forecasting under multiple bidding types, giving prominence to the difficulty of a unified forecast service. First, despite the historical auction logs, Campaign Performance Forecasting bears the brunt of obtaining response information on retrieved historical impressions for advertisers in time, which is especially essential for the $replay$ idea to work from CPM to CPC. Furthermore, performance forecasting on automatic bidding(BCB, MCB) has never been studied, leaving an almost blank area for us.  

Moreover, as shown in Figure \ref{auction_process}, most advertising platforms apply complex strategies to optimize online ads delivery, like parallel real-time retrieval\cite{zhu2019joint}, bidding strategies\cite{zhu2017optimized,wu2018budget,yang2019bid} and user response calibration\cite{pan2020field}. Assuming that every segment in bidding system can be reproduced exactly regardless of any cost, then we can replay on historical logs to acquire accurate forecast results for campaigns. However, these strategies, which usually bring great fluctuation on campaign performance, are hard to reproduce entirely under limited resource and low latency, thus serious deviation may occur in replay process. Taking Figure \ref{auction_process} as an example, because of the hard-to-reproduce real-time retrieval strategy in match phase, some audience could be missed in replay and thus resulting in an underestimated performance.

To sum up, although the $replay$ idea ensures the $interpretability$ on forecast results naturally, Campaign Performance Forecasting faces two more challenges for real-world applications. (1) How to estimate campaign performance for various bidding types in a unified replay framework. (2) How to mitigate the deviation of estimation brought by hard-to-reproduce strategies in replay process, providing more accurate forecast results. To the best of our knowledge, existing works have not studied these challenges yet.  

In this paper, we propose a novel framework for Campaign Performance Forecasting, which mainly consist of \textit{performance estimation} and \textit{performance calibration}. In \textit{performance estimation}, an user response forecasting module is deployed to predict interactive probabilities between online requests and advertisers continuously, completing the auction records for replay. After that we formulate a unified replay algorithm for various bidding types, in which match and rank phase are reproduced on historical logs to estimate campaign performance, ensuring the interpretability on results. In \textit{performance calibration}, inspired by the success of multi-task learning\cite{caruana1997multitask,jacobs1991adaptive,ma2018modeling} and its wide applications\cite{ren2015faster,eigen2013learning,shazeer2017outrageously} in DNNs, we firstly analyze the Pearson Correlation among forecast tasks, then we innovatively propose a multi-task learning method to calibrate the deviation brought by hard-to-reproduce strategies in \textit{performance estimation}. The proposed method captures mixture calibration patterns of task assembling, learns the calibration function to map the estimated performance to the true ones. In summary, \textit{performance estimation} reproduces essential bidding process on auction logs, ensuring the $interpretability$ on results, \textit{performance calibration} calibrates the replay deviation to achieve a better accuracy, thus our framework could offer advertisers interpretable and accurate forecast results under multiple bidding types.



We conduct our experiments on a real world dataset from Taobao .com. The experiment results demonstrate that our proposed framework outperforms other baselines by a large margin, and it shows an excellent robustness for adapting volatile bidding environment. An online A/B test deployed on Taobao display advertising platform yields a $8\%$ $arpu$ lift, demonstrating the effectiveness of our method in the real world. In addition, our model is more efficient than the models built separately since it models multiple forecast targets in an multi-task fashion.


Our contributions in this paper are three-fold:  (1) A unified framework for Campaign Performance Forecasting, which is capable of providing insight for advertisers under various bidding types, and it can be well generalized to any advertising platform. (2) Empirical verification of the proposed framework on a real-world dataset from Taobao.com, and practical service deployment on Taobao display advertising platform, meeting both accuracy and interpretability requirements. (3) A benchmark dataset for public, which fertilizes the research of Campaign Performance Forecasting. To the best of our knowledge, it's the first dataset for this very task. 

In the remainder of this paper, we will first review the related work in Section 2, then we elaborate on the proposed method in Section 3. After that, we introduce the experiment settings in Section 4, results and discussion follow in Section 5. Finally, the conclusion and future work in Section 6.


\section{Related Work}
In this paper, we propose a unified framework to forecast campaign performance under various bidding types. To the best of our knowledge, few prior works focus on this very task. Based on the overall content of this paper, we review the most related domains here, user response forecasting and campaign performance forecasting.

\subsection{User Response Forecasting}
User response forecasting\cite{zhou2018deep,zhou2019deep,ma2018entire,lee2012estimating,guo2017deepfm} is a popular area in online advertising, which is usually formulated as a binary classification problem to learn the Click-Through Rate(CTR) or Conversion Rate(CVR) at <user, creative> level. User response forecasting in Campaign Performance Forecasting is a little different from above works. Since advertisers usually upload totally new creatives when planning a new campaign, we formulate a model to predict the CTR/CVR at <user, advertiser> level in consideration of the better stability on 'advertiser' than 'creative', ensuring response information on auction logs could be prepared for replay in advance. The model takes DeepFM\cite{guo2017deepfm} as backbone, which has been proven lightweight but effective in binary classification tasks. Differently, inheriting the advantages of DeepFM for learning low- and high-order interaction among sparse features, we extra introduce interactive behaviors between users and advertisers into model as dense features to bring additional improvements.


\subsection{Campaign Performance Forecasting}
As for Campaign Performance Forecasting, the most related works focus on forecasting granular audience size for DSPs\cite{yuan2013real}, the problem is claimed to be tough because of the large number of combinations among campaign criteria, which increases explosively with an exponent. Shi et al.\cite{shi2018audience} takes it as a data compression problem. They propose a MinHash method to mine granular audience size from historical data. Similarly, a Frequent Item set Mining(FIM) algorithm \cite{sinha2019forecasting} has been proposed to mine historical audience size for frequent criteria combinations, and then estimate its future audience with time-series forecasting models. 

Contrary to academic work, Campaign Performance Forecasting has attracted sufficient attention in industry because of the business value it brings. Most global adverting platforms like Google, Facebook and Taobao offer analogous forecast tools for campaign optimization. Many patents\cite{cui2013campaign,kalish2016method,wang2012forecasting,jiang2015predicting} have been applied by these platforms to claim their ideas. Kalish et al.\cite{kalish2016method} proposes a framework that retrieves similar existing campaigns to assist performance forecasting for new campaigns. However, campaigns from different advertisers may get much different performance even with the same criteria. Jiang et al.\cite{jiang2015predicting} proposed an replay idea which reproduces match and rank process on historical auction logs to observe campaign performance. However, the method works on CPM bidding only, and it fails to forecast winning clicks or conversions for advertisers.

Our work is a replay based approach since its natural interpretability on forecast results. The main difference is that we consider to estimate campaign performance for multiple common bidding types in a unified framework. Moreover, to mitigate the estimation deviation brought by strategies that's hard to replay on historical logs, we innovatively introduce multi-task learning\cite{caruana1997multitask,jacobs1991adaptive,ma2018modeling} to calibrate the estimated results to the true ones in this paper. Although wide applications for multi-task learning in DNNs\cite{ren2015faster,eigen2013learning,shazeer2017outrageously}, our work is the first to apply it in Campaign Performance Forecasting for calibration.

\section{Methodology}
\subsection{Preliminaries}
As is well known, advertisers usually have a flexibility for combining various criteria when planning campaigns for online display advertising, we present the most common criteria as follows.
\begin{itemize}[leftmargin=*]
\item \textbf{Hour.} Advertisers could choose when their ads deliver. e.g. Delivery starts at 7:00 a.m and ends at 23:00 p.m everyday.
\item \textbf{Geographical Area.}  It is usually formatted as combinations of countries, states or provinces etc. Only the impression opportunities from chosen areas can be reached by the ads.

\item \textbf{Adzone.} Most global advertising platforms provide more than one adzone for choice, advertisers usually choose adzones in consideration of its online traffic scale, and provide creatives with suitable size.

\item \textbf{Targeting option.} Getting to targeting audience is a crucial step for a successful advertising campaign. For instance, 1) Targeting people with specific profiles like age, gender, used devices, income level, etc. 2) Targeting past users who have engaged with your ads. 3) Targeting the users who are similar to your existing audience, or the audience of another advertiser who is similar to you. Targeting option usually consists of id tags.

\item \textbf{Objective.} Marketing goal of an advertising campaign, e.g. maximum clicks, maximum conversions. 

\item \textbf{Budget.} Budget is the money advertisers are willing to spend on presenting ads to users, it is usually used for cost control. 

\item \textbf{Bidding Type.} Bidding types provided by advertising platforms to meet personalized delivery. Manual bidding like CPM, CPC, CPA and automatic bidding like BCB, MCB are the most common ones. For manual bidding, advertiser should provide an fixed bid price, which usually set with Cost-Per-Mille, Cost-Per-Click or Cost-Per-Action. For automatic bidding, advertiser should set a price for unit cost constraint in MCB.

\end{itemize}

With all these components, Campaign Performance Forecasting service should return the estimated key performance indicators in seconds after the forecasting interface receiving above specific parameters, helping advertisers to optimize their advertising campaigns.

\subsection{Problem Formulation}
Given a new advertising campaign $\mathcal{C}$ with criteria set $\mathcal{S}$, the performance is defined as $\#impression$, $\#cost$ and $\#click$ it will acquire in $T$-th day. Suppose that we have a historical auction set $\mathcal{D}=\{X_i\}_{i=1}^{N}$ from $T$$-$$1$-th day, where $N$ is number of auctions, which rarely fluctuates from day to day. $X_i$ is an detailed auction record, including the winner, highest bid price, predicted CTR etc. Our work is to build a unified framework $\mathcal{F}$ that forecasts performance for campaign $\mathcal{C}$, which is formulated as $\#impression$,$\#cost$,$\#click$$=$$\mathcal{F}(\mathcal{S}, \mathcal{D})$.

\subsection{Campaign Performance Forecasting}
Campaign Performance Forecasting should work on multiple bidding types and provide interpretable and accurate estimation indicators for advertisers in seconds. In this section, we focus on presenting how to construct a unified forecast framework to achieve above goals. The framework mainly consists of two parts, performance estimation and performance calibration. Unified replay in performance estimation ensures the interpretability on forecast results under multiple bidding types, and the latter calibration module mitigates the replay deviation for better accuracy. Thus, the combination of $replay$ and $calibration$ could produce interpretable and accurate forecast, while a regular regression model is hard to maintain interpretability on any campaign criteria dimension.

\begin{figure}[H]
\centering 
\includegraphics[width=0.485\textwidth]{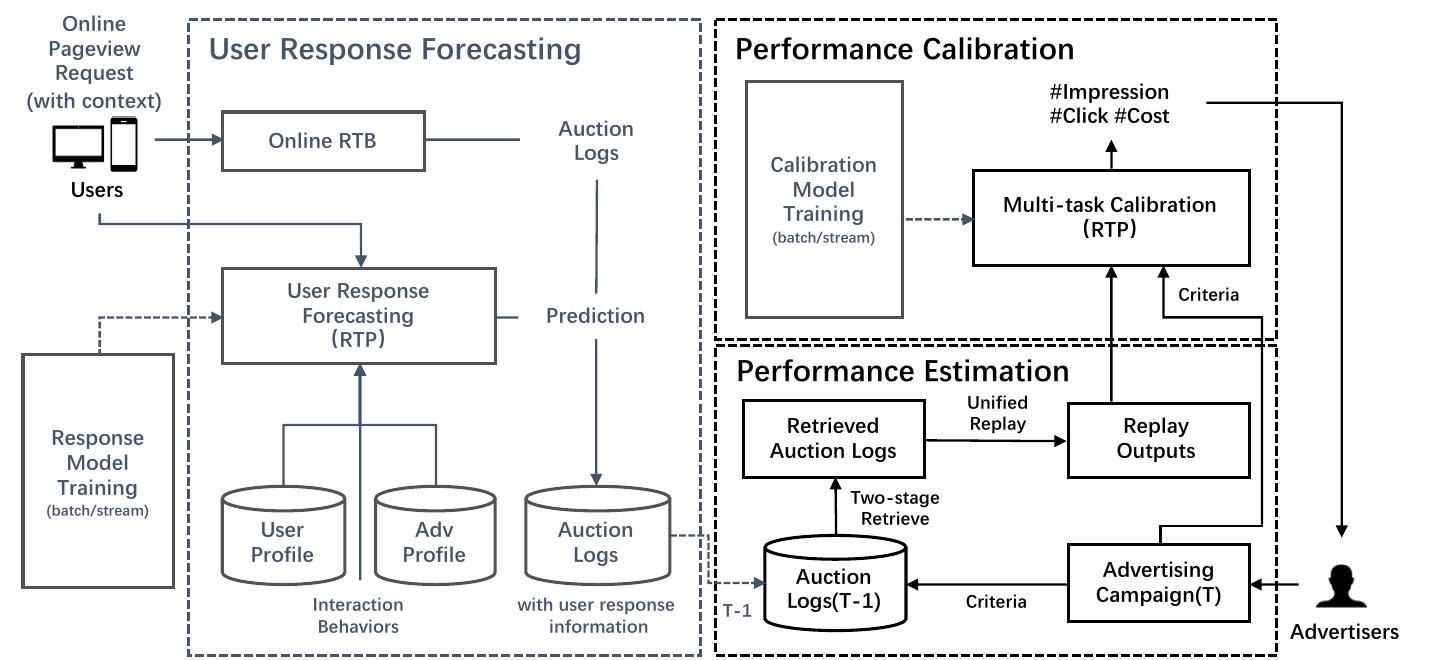}
\caption{The overall framework of Campaign Performance Forecasting.}
\label{campaign_performance_forecasting} 
\end{figure}

Specifically, in performance estimation, a unified replay algorithm is employed to estimate campaign performance under various bidding types, in which only essential match and rank process in bidding system are reproduced for low computation complexity. In performance calibration, we innovatively propose a multi-task calibration method which learns correlation between forecast indicators to map the esitimated results to the true ones, improving both accuracy and efficiency. For better display of the proposed method, the overall framework is illustrated in Figure \ref{campaign_performance_forecasting} , and the detailed process is described in following sections. We present more reproducibility details in supplementary material, including the construction of base log, unified replay algorithm, calibration model, and online latency optimization. The code and data will be released at {\color{blue} \url{https://github.com/anonymousauthorss/reproducibility}}.

\subsubsection{User Response Forecasting}
\ 
\newline 
User response information like CTR/CVR is critical for replay in consideration of its indispensability in auction process. To ensure user response information of historical requests can be fetched under any targeting criteria made by advertisers, an online forecasting module is deployed to predict CTR/CVR of online pageview requests on advertisers continuously.


As for the forecasting model, we all know that the more complex the model is, the more accurate the CTR/CVR information for replay will be. In our design, any effective model could be adopted as backbone in user response forecasting module. For instance, if the platform owns sufficient computing resources, a way to achieve the most accurate response results is to reproduce online deployed Real-Time-prediction(RTP) models, like DIEN\cite{zhou2019deep} in Taobao.com. In contrary, a lightweight but effective model like DeepFM\cite{guo2017deepfm} also could be a good choice under insufficient computing resources, and the forecasting deviation brought by suboptimal model will be handled well in the following calibration module. In summary, a wide selection on model ensures that user response forecasting module could be generalized to any other advertising platform.

For this paper, we employ the lightweight DeepFM as backbone, more details about the model can be seen in \cite{guo2017deepfm}. Besides sparse features like profile and context information, interactive behaviors like browse and buy between users and advertisers are introduced into the model as dense features, which explicitly models user interest to improve model performance. According to our experiment results, the model achieves an $\bm{AUC}$ of $\bm{0.88}$ and an $\bm{LogLoss}$ of $\bm{0.057}$ for CTR forecasting, similar results occurs in CVR forecasting.







With user response information produced by the module, complete auction records (including both auction and user response information) on historical requests could be retrieved for a campaign in time, laying the foundation of performance estimation.

\subsubsection{Performance Estimation}
\ 
\newline 
In this part, we elaborate how we estimate campaign performance under various bidding types in a unified replay algorithm. Without loss of generality, we first review the bidding process under second price auction mechanism\cite{edelman2007internet} when advertisers participates in auctions.

Suppose that there are $N$ impression opportunities arriving sequentially ordered by an index $i$ in a day. The advertiser provides a bid $b_i$ and competes with other bidders in real-time. If $b_i$ is the highest bid in the auction, the advertiser wins the impression $i$ with a price charged $c_i$, which equals to the highest bid price of other bidders. The bidding process terminates whenever delivery results have reached the limits set by advertisers, for example, the total cost reaches the $budget$, or all the impression opportunities have gone through the auction. $b_i$ here is a common currency for comparison when ads with different bidding types are competing for a same opportunity.

\textbf{Manual bidding.} For Cost-Per-Mille(CPM), Cost-Per-Click(CPC) and Cost-Per-Action(CPA) bidding, advertisers give a fixed $bidprice$ to participate auctions, expecting to win impressions in which the corresponding $b_i$ is the highest bid. It emphases on cost control since the price charged for every winning impression is less than $b_i$. $b_i$ is calculated from $bidprice$ offered by advertisers as follows.

\begin{equation}
 b_i=\left\{
\begin{array}{lcl}
bidprice & & {CPM }\\
bidprice \times pctr_i \times 1000 & & {CPC}\\
bidprice \times pctr_i \times pcvr_i \times 1000 & & {CPA}
\end{array} \right. 
\end{equation}
where $pctr_i$, $pcvr_i$ is the predicted CTR/CVR of impression $i$ on advertiser from the user response forecasting module.

Apparently, performance estimation for manual bidding campaigns can be determined by compare its bid $b_i$ to the highest bid price recorded in retrieved historical auctions. 

\textbf{Automatic bidding.} For Budget-Constrained-Bidding(BCB)\cite{wu2018budget}, advertisers hope to gain as much the total value of winning impressions as possible under a limited $budget$. Let $x_i$ be a binary indicator whether the advertiser wins impression $i$, and $v_i$ represents the value of impression $i$, the goal of budget constrained bidding is formulated as:

\begin{equation}
\begin{aligned}
    maximize \sum_{i=1}^{N}{x_iv_i}\\
    s.t. \sum_{i=1}^{N}{x_ic_i} <= budget
\end{aligned}
\end{equation}

Despite the maximal total value budget constrained bidding brings, the average cost per obtaining is not controllable theoretically, while it is routinely one of the most concerned indicator for advertisers. Mutivariable-Constrained-Bidding(MCB)\cite{yang2019bid} is a complement to this drawback, which extends budget constrained bidding by providing an additional $constraint$ for unit cost in bidding strategy. Advertisers usually set a corresponding $constraint$ according to their marketing goals. For example, an advertiser wants to obtain a maximal click, and the Pay-Per-Click(PPC) don't exceed \$2. The formulation of mutivariable constrained bidding is shown in Eqn.(\ref{MCB}).

\begin{equation}
\begin{aligned}
    maximize & \sum_{i=1}^{N}{x_iv_i}\\
    s.t. \sum_{i=1}^{N}{x_ic_i} &<= budget \\
    \frac{\sum_{i=1}^{N}{x_ic_i}}{\sum_{i=1}^{N}{x_iv_i}} &<= constraint
\end{aligned}
\label{MCB}
\end{equation}
where $v_i$ is an expression of objectives set by advertisers, see Eqn.(\ref{VI}). Notably, we replace the real occurrence with predicted probabilities in this section, which facilitates a more concise formulation and theoretical analysis, with a trivial influence in practice.

\begin{equation}
v_i = \left\{
\begin{array}{lcl}
1 & & {Obj.~Impression}\\
pctr_i & & {Obj.~Click}\\
pctr_i \times pcvr_i & & {Obj.~Conversion}\\
\end{array} \right. \\
\label{VI}
\end{equation}

As described in \cite{wu2018budget,yang2019bid}, $b_i$ for constrained bidding is given by bidding agents according to the impression value $v_i$, and the optimal bidding strategies are formalized as follows, where $\lambda, \alpha, \beta$ are scaling factors.

\begin{equation}
 b_i=\left\{
\begin{array}{lcl}
\lambda \times v_i & {BCB}\\
\alpha \times v_i + \beta \times v_i \times constraint & {MCB}\\
\end{array} \right. \\
\end{equation}

The better bidding strategies perform, the closer the campaign performance is to the optimal\footnote{It is usually measured with $R/R^*$, which approaches $1$ as described in \cite{wu2018budget,yang2019bid}.}. Therefore, optimal performance could be a approximate reference to help advertisers optimize their advertising campaigns. Budget constrained bidding is formalized to a knapsack problem by Lin et al.\cite{lin2016combining}, in which the optimal performance can be derived through greedy approximation algorithm \cite{zhang2014optimal,zhang2016optimal} when the impression opportunity set is known. Therefore, performance for budget constrained bidding campaigns can be obtained given a known auction set with the following steps. (1) Sort the 
retrieved impressions in auction set by $\frac{c_i}{v_i}$; (2) Choose impressions from the sorted set continuously until the total $\sum_i{c_i}$ reaches the $budget$; (3) Observe campaign performance from the chosen impressions. Performance estimation for multivariable constrained bidding is highly analogous since its bid formula could be reorganized to $b_i = v_i\times ({\alpha+\beta \times constraint})$. The only difference is a additional condition, whether $\frac{\sum_{i}c_i}{\sum_{i}{v_i}}$ reaches the $constraint$.

\begin{algorithm}[!htb]
\caption{The Unified Replay Algorithm}
\label{alg:PerformanceEstimate}
\LinesNumbered 
\KwIn{historical auction log $\mathcal{D}$, campaign criteria set $\mathcal{S}$ from advertiser $adv$.}
\KwOut{estimated $\#impression$, $\#cost$ and $\#click$.}
init $\#impression$, $\#cost$, $\#click$ with $0$\;

//Match Phase\;
two-stage retrieval subset $\mathcal{D'}$ from $\mathcal{D}$ for the campaign\;
\For{$X_i$=$\{user_i, hour_i, pctr_i, v_i, b^1_i, b^2_i, winner_i\}$ $in$ $\mathcal{D'}$}{
    $c_i$=$b^1_i$;//$c_i$ represents the cost to win the $i$-th auction\;
    \If {$winner_i$=$adv$}{
    $c_i$=$b^2_i$\;
    }
    append $c_i$ to set $X_i$\;
}
//Rank Phase\;
\If {$bidding\_type = BCB~\|~MCB$}{
init $\#value, condition$ to $0, True$ respectively\;
sort $\mathcal{D'}$ by $\frac{c_i}{v_i}$\;
\For {$Y_i$=$\{user_i, pctr_i, v_i, c_i\}$ $in$ $\mathcal{D'}$ }{
\If {$bidding\_type = MCB$}{
$condition = \frac{\#cost}{\#value}<constraint$\;}
\If {$\#cost$<$budget~\&~condition$}{
add $1$, $c_i$, $pctr_i$ to $\#impression$, $\#cost$, $\#click$\;
add $v_i$ to $\#value$\;
}
}
}
\ElseIf{$bidding\_type = CPM~\|~CPC~\|~CPA$}{
$b$ = $bidprice$\;
sort $\mathcal{D'}$ by $hour_i$\;
\For {$Y_i$=$\{user_i, pctr_i, v_i, c_i\}$ $in$ $\mathcal{D'}$ }{
$b$ = $bidprice*v_i*1000$\;
\If {$\#cost$<$budget~\&~b$>$c_i$}{
add $1$, $c_i$, $pctr_i$ to $\#impression$, $\#cost$, $\#click$\;
}
}
}
\If {$\#cost$>$budget$}{
times $\frac{budget}{\#cost}$ to $\#impression$, $\#cost$, $\#click$\;
}
\Return{$\#impression$, $\#cost$ and $\#click$.}
\end{algorithm}

To further describe the estimation algorithm, we briefly explain the core symbols mentioned in this section as follows.

\begin{itemize}[leftmargin=*]
    \item {$\bm{\mathcal{D}}$:} Auction records with complete information from $T$$-$$1$-th day, donated as $\mathcal{D}=\sum_{i=1}^{N}X_i$. Record $X_i$ is formulated as $(user_i, tag_i, v_i,$\\$pctr_i, hour_i, area_i, adzone_i, winner_i, b_i^1, b_i^2)$, where $user_i, hour_i,$\\$area_i, adzone_i$ are the context of $i$-th page request, $b^1_i, b^2_i, winner_i$ are the highest, second highest ecpm bid price and the winner of the $i$-th auction respectively, $tag_i$ is targeting tags on $user_i$. $\mathcal{D'}$ is a subset of $\mathcal{D}$, which contains auction records of a specific campaign after the two-stage retrieval\cite{li2021truncationfree} by the "$campaign$$-$$>$$target$\\$ing\_option\_tags$ + $tags$$-$$>$$user$ = $campaign$$-$$>$$user$" rule in match phase.
    \item {$\bm{\mathcal{S}}$:} Campaign criteria set $S$ from an advertising campaign made by advertisers at $T$-th day. As described in Section 3.1, it usually includes $hours, areas, adzones, targeting\_options, bidding\_type, $\\$ budget, objective$, and a $bidprice$ for manual bidding, a $constraint$ for mutivariable-constrained bidding.
\end{itemize}

\textbf{Latency optimization.}
Most of the computation concentrates in Algorithm \ref{alg:PerformanceEstimate}, we adopt several speed-up actions to optimize the latency for online service. (1) Down sampling is applied in auction logs stream to reduce computation complexity. (2) Only essential Match Phase and Rank Phase in auction process are considered in replay, neglecting complex strategies like parallel retrieval\cite{zhu2019joint} which are hard to reproduce. 

So far, Algorithm \ref{alg:PerformanceEstimate} could provide interpretable forecast results under multiple bidding types, it is not good enough for direct service for advertisers in consideration of the replay deviation brought by neglected strategies. Next, we will introduce our calibration module, which calibrates the deviation in replay to achieve more accurate forecast results.

\subsubsection{Performance Calibration}
\ 
\newline 
Deviation between the estimated performance and the true ones need to be calibrated for real-world applications. Obviously, there are two modes for calibration methods, one by one or all at the same time. For Campaign Performance Forecasting, the service is useful only when it responds to advertiser requests fast, and the Pearson Correlation analysis following Ma et al. \cite{ma2018modeling} shows a high relevance among $\#impression,\#cost,\#click$ either (see Table \ref{tab:pearson}). Therefore, we innovatively introduce multi-task learning in Campaign Performance Forecasting to calibrate estimation deviation, giving consideration to both efficiency and accuracy improvement as we discussed above. 

\begin{figure}[!htp]
\centering 
\includegraphics[width=0.5\textwidth]{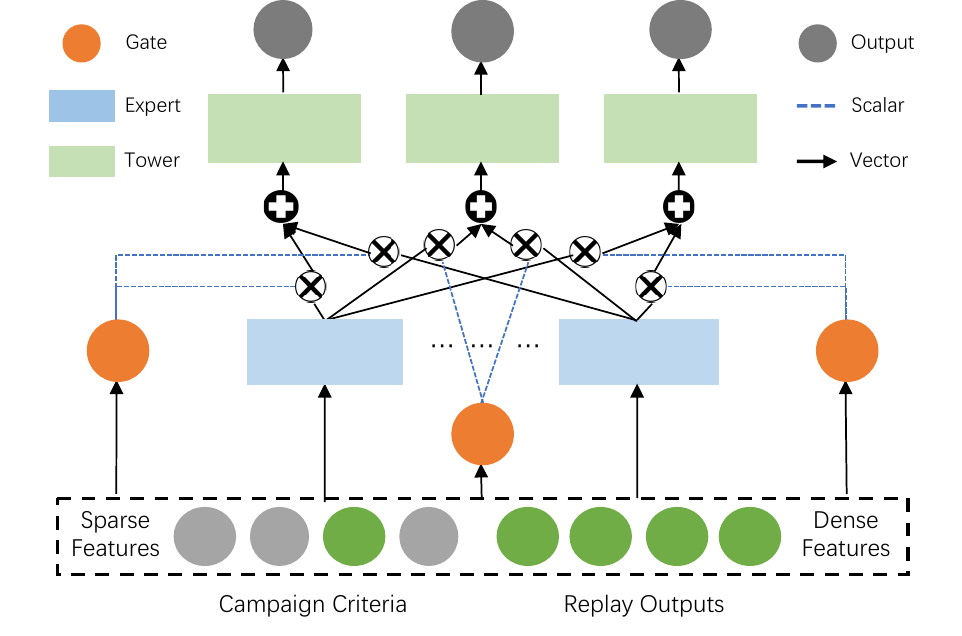}
\caption{Architecture of the Multi-task Calibration model.}
\label{muti_task_calibration} 
\end{figure}

The model structure is shown in Figure \ref{muti_task_calibration}, the $expert$ is a feed-forward network, and the $gate$ takes the input features and outputs softmax probabilities to assemble the experts with different weights, allowing different tasks to obtain different information from experts. Then, the assembled results are feed into the task-specific $tower$ networks for calibration function learning. Specifically, the formalization is as follows.
\begin{equation}
\begin{aligned}
     y^k &= h^k(f^k(x))\\
     f^k(x) &= \sum_{i=1}^{N}g^k(x)_if_i(x)\\
     g^k(x) &= softmax(W_{gk}x)
\end{aligned}
\end{equation}
where $x$ is the input, $y^k$ is the calibrated output of $k$-th task. $f^k$ and $g^k$ are the output of $k$-th $expert$ and $gate$ respectively. $W_{gk} \in R^{N \times D}$ is trainable parameters, where $D$ is the dimension of input features, $N$ is the number of $expert$.

More specifically, the input consists of two parts, among that the replay outputs act as base performance, and campaign criteria is adopted into model for calibration patterns capturing. For example, a campaign chooses a targeting option with parallel real-time retrieval, then it usually has a better true performance than the estimated base because of the audience loss from two-stage retrieval strategy in Algorithm \ref{alg:PerformanceEstimate}, thus the calibration function should increase all the estimated results appropriately. The input are then fed into multiple expert networks, which are shared across all calibration tasks. The gating networks for different tasks can learn different mixture patterns of experts assembling to optimize calibration performance by capturing the task relationships. Notably, the model is learned to map estimated results to the true ones in an multi-task fashion, which eliminates heavy labor on maintaining multiple forecast models.

\begin{table}[!htb]
\setlength{\abovecaptionskip}{3pt}
\setlength{\belowcaptionskip}{10pt}
\caption{Key components of the calibration dataset.}
\label{table:dataset}
\centering
\footnotesize 
\resizebox{\linewidth}{!}{%
\begin{tabular}{lp{6.0cm}}
\hline 
\textbf{Domain} & \textbf{Description} \\
\hline
Input & Campaign criteria as described in Section 3.1, statistical features of match phase and final results from Algorithm \ref{alg:PerformanceEstimate}.\\ \hline
Labels &  Observed true performance of a campaign.\\
\hline
Quantity & The manual bidding set contains $6899/71298$ CPM/CPC campaigns, among that $74175$ for train and $4022$ for evaluation. The automatic bidding set contains $23313$ BCB campaigns, $22238$ for train and $1075$ for evaluation.\\
\hline
\end{tabular}
}
\end{table}

\section{Experiment Settings}
\subsection{Dataset}
It costs not so much effort to construct logs for user response forecasting and performance estimation on Taobao.com when the log server records the whole log stream. However, collecting data for calibration model training and evaluating is nontrivial because of the frequent adjustments on campaigns, which leads to volatile campaign performance in a day. We observe campaigns on Taobao.com for a long time from 01/15/2020 to 05/10/2020, and collect the ones with no pause and no criteria adjustment to construct our calibration dataset. The key components of the dataset could be summarized as Table \ref{table:dataset}.

\subsection{Evaluation Metrics}
A cost-weighted $\bm{mape}$\cite{de2016mean} is employed to evaluate the accuracy for all methods according to the business background, and a $\bm{ratio_p}$ is adopted to measure the the distribution of their forecasting results. A widely used $\bm{arpu}$\cite{mccloughan2006accounting} is employed for online A/B test to analyze the revenue generation capability and growth at advertiser-level brought by our method.

\begin{equation*}
\begin{aligned}
   ape_i &= \frac{|y^{'}_{i}-y^{*}_{i}|}{|y^{*}_{i}|}\\
   mape &= \frac{\sum_i{(cost_i * ape_i)}}{\sum_i{cost_i}}\\
   ratio_p &= \frac{\sum_i{I_{ape_i<p}}}{\sum_i{1}}
\end{aligned}
\end{equation*}
where $y^{'}_{i}$, $y^{*}_{i}$ represent the forecasting and true performance respectively. $cost_i$ is the true cost of a campaign. $p$ is a given constant $0.5$, and $I$ is a binary indicator function.


\subsection{Compared Methods}
In view of few related work in Campaign Performance Forecasting, we conduct a series of experiments with the proposed approach, and several strong baselines are adopted for comparison. All the methods are described as bellow.

\begin{itemize}[leftmargin=*]
    \item {$\bm{MTL_N}$.} Our proposed method in this paper, which first replays campaign performance on historical auctions as described in Algorithm \ref{alg:PerformanceEstimate}, and then calibrates the base replay results in a multi-task learning architecture.
    \item {$\bm{REPLAY}$.} Our proposed unified replay method in Algorithm \ref{alg:PerformanceEstimate}, which reproduces auction process on historical logs to observe campaign performance under both manual bidding and automatic bidding. Notably, our method naturally surpasses the approach proposed by Facebook\cite{jiang2015predicting}, which only works on manual bidding.
    \item {$\bm{GBDT_1}$.} The method adopts the same input as ${MTL_N}$, and employs a GBDT\cite{ke2017lightgbm} model to calibrate the replay results from Algorithm \ref{alg:PerformanceEstimate} one by one.
    \item {$\bm{MTL_1}$.} The method adopts the same input and neural architecture as ${MTL_N}$, and it calibrates the replay results from Algorithm \ref{alg:PerformanceEstimate} one by one.

\end{itemize}

With all these experiments, we could  better elaborate the effectiveness of our proposed framework by the following comparisons. (1) To prove the necessity of calibration by comparing the calibration methods to the replay methods. (2) To verify the effectiveness of neural architecture of multi-task learning by comparing ${MTL_1}$ to ${GBDT_1}$. (3) To demonstrate the improvement brought by multi-task learning by comparing ${MTL_N}$ to ${MTL_1}$.

\section{Results and Analysis}
\subsection{Correlation Analysis}
Prior works\cite{ma2018modeling,misra2016cross} indicate that the performance of multi-task learning models highly depends on the inherent task relatedness in the data. Follow the description in \cite{ma2018modeling}, we adopt Pearson Correlation among the labels of tasks as the quantitative indicator of task relationships. We measure the task correlation among $\#impresion$, $\#cost$ and $\#click$ on the real-world dataset from Taobao.com. The correlation analysis results are shown in Table \ref{tab:pearson}, it is clear that the forecast tasks show a high relevance between each other, that's a positive signal for the effectiveness of multi-task learning.

\vspace{-5pt}
\renewcommand{\arraystretch}{1.5}
\begin{table}[!htp]
\setlength{\abovecaptionskip}{3pt}
\setlength{\belowcaptionskip}{10pt}
  \centering
  \tiny 
  \caption{Pearson correlation among forecasting indicators.}
  \label{tab:pearson}
    \resizebox{\linewidth}{!}{%
    \begin{tabular}{c|c|c|c}
    \hline
    \multirow{2}{*}{\textbf{Option}}&
    \multicolumn{3}{c}{\textbf{Pearson Correlation}}\cr\cline{2-4}
    &$\bm{\#impression}$ $\sim$ $\bm{\#click}$ &$\bm{\#impression}$ $\sim$ $\bm{\#cost}$ 
    &$\bm{\#cost}$ $\sim$ $\bm{\#click}$ \cr
    \hline
    \hline
    Manual bidding&0.76&0.79&0.81\cr
    \hline
    Automatic bidding&0.80&0.72&0.74\cr
    \hline
    \end{tabular}
    }
\end{table}

\vspace{-5pt}
\subsection{Experimental Results}
The performance for Campaign Performance Forecasting of different methods on the Taobao dataset is shown in Table \ref{tab:performance}, the numbers in bold and underlined represent the best and the second best performance respectively. 

In terms of manual bidding, it's clear that $REPLAY$ performs poorly on all indicators, especially on $mape$. The method suffers from serious deviation brought by the hard-to-reproduce strategies in replay. On contrary, the calibration methods $GBDT_1$, $MTL_1$ and $MTL_N$ outperform the $REPLAY$ baseline by a large margin on $mape$, indicating the effectiveness of calibration on this very task. More specifically, despite sharing the same calibration mode, $MTL_1$ surpasses $GBDT_1$ on most indicators, demonstrating the effectiveness of the base architecture from $MTL$. As we consider the correlation among multiple calibration tasks, the overall results of $MTL_N$ are better than $MTL_1$ significantly (t$-$test with $p < 0.01$). This is mainly because $MTL_N$ explicitly models the task relationships and learns different mixture patterns of task assembling to improve performance for all tasks. Regarding $ratio_{0.5}$, $MTL_N$ surpasses other methods, demonstrating that multi-task learning not only can calibrate the outliers to improve the overall $mape$, but also make better distribution with more accurate forecasting results appearing in the $ape<0.5$ range.

\vspace{-8pt}
\newcommand{\tabincell}[2]{\begin{tabular}{@{}#1@{}}#2\end{tabular}}  
\renewcommand{\arraystretch}{1.5}
\begin{table}[!htp]
\setlength{\abovecaptionskip}{3pt}
\setlength{\belowcaptionskip}{10pt}
  \centering
  \tiny 
  \caption{Experiment results on Campaign Performance Forecasting.}
  \label{tab:performance}
    \resizebox{\linewidth}{!}{%
    \begin{tabular}{c|c|c|c|c|c|c|c}
    \hline
    \multirow{2}{*}{\textbf{Option}}&
    \multirow{2}{*}{\textbf{Method}}&
    \multicolumn{2}{c|}{$\bm{\#impression}$}&
    \multicolumn{2}{c|}{$\bm{\#click}$}&
    \multicolumn{2}{c}{$\bm{\#cost}$}\cr\cline{3-8}
    & 
    &$\bm{mape}$ & $\bm{ratio_{0.5}}$
    &$\bm{mape}$ & $\bm{ratio_{0.5}}$
    &$\bm{mape}$ & $\bm{ratio_{0.5}}$\cr
    \hline
    \hline
    \multirow{4}{*}{\tabincell{c}{Manual \\bidding}}
    & $REPLAY$ & 4.34&24\%&2.96&21\%&2.27&22\%\cr\cline{2-8}
    & $GBDT_1$ & 1.13&28\%&1.18&25\%&1.13&28\%\cr\cline{2-8}
    & $MTL_1$ & \underline{0.82}&\underline{34\%}&\underline{1.02}&\underline{25\%}&\underline{0.89}&\underline{28\%}\cr\cline{2-8}
    & $MTL_N$ & \textbf{0.78}&\textbf{37\%}&\textbf{0.90}&\textbf{32\%}&\textbf{0.83}&\textbf{33\%}\cr\cline{2-8}
    \hline
    \hline
    \multirow{4}{*}{\tabincell{c}{Automatic \\bidding}}
    & $REPLAY$ & 2.49&18\%&3.31&16\%&\textbf{0.15}&\textbf{89\%}\cr\cline{2-8}
    & $GBDT_1$ & 0.87&\underline{44\%}&1.33&26\%& / & / \cr\cline{2-8}
    & $MTL_1$ & \underline{0.86}&42\%&\underline{1.18}&\underline{32\%}& / & / \cr\cline{2-8}
    & $MTL_N$ & \textbf{0.85}&\textbf{47\%}&\textbf{0.99}&\textbf{34\%}& / & / \cr\cline{2-8}
    \hline
    \end{tabular}
    }
\end{table}

As for automatic bidding, evaluation performance on $cost$ is extremely high because BCB campaigns always spend its budget all, thus no calibration needed. Regarding other tasks, we can see that $MTL_N$ achieves much better improvement on $impression$ and $click$ than other methods, that is consistent with the performance on manual bidding. It indicates that our proposed method is universal on multiple bidding types.

\begin{figure}[!htp]
\centering
\subfigure[Manual bidding.]{
\begin{minipage}[t]{0.5\linewidth}
\centering
\includegraphics[width=1.6in]{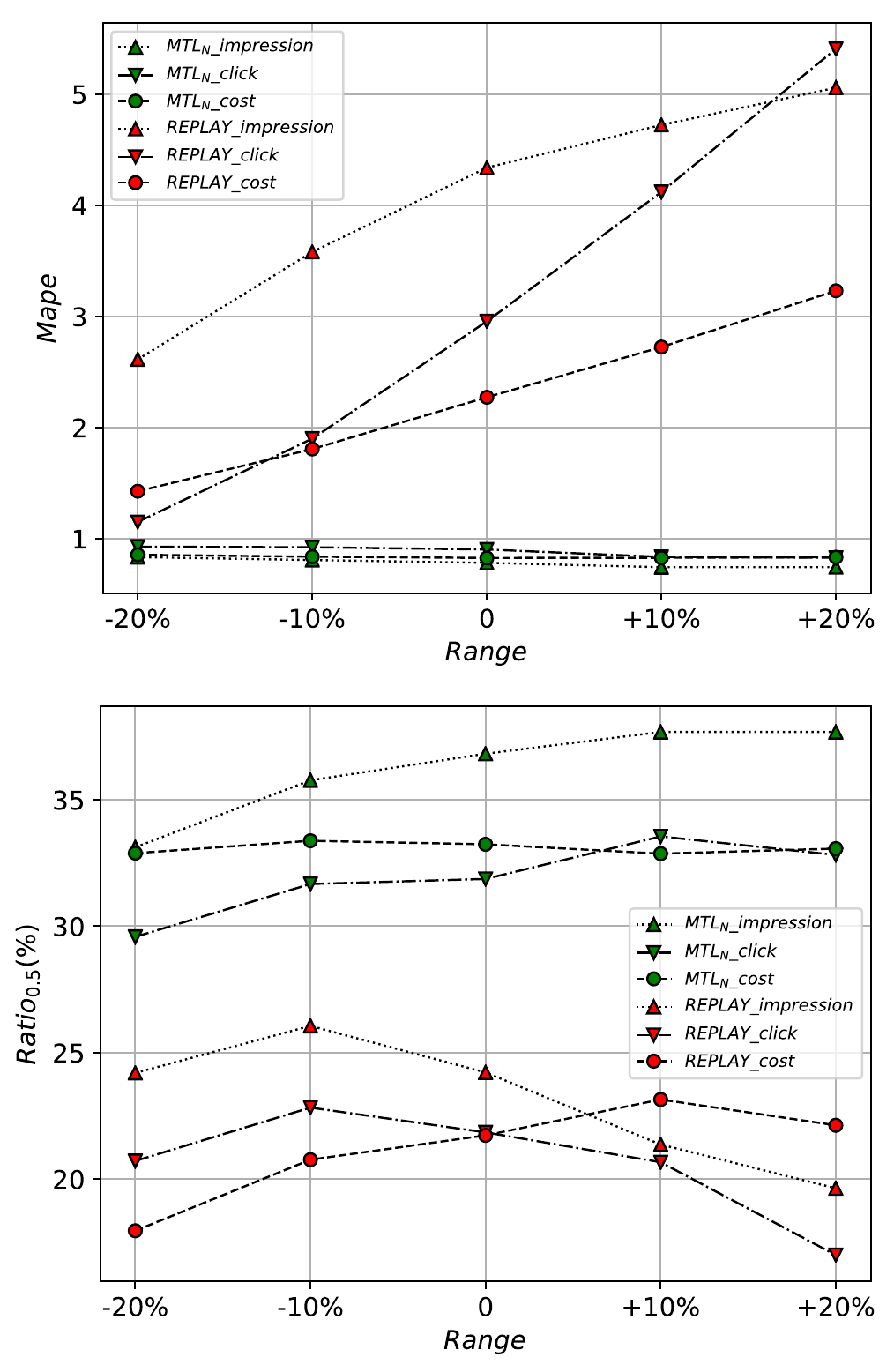}
\end{minipage}%
}%
\subfigure[Automatic bidding.]{
\begin{minipage}[t]{0.5\linewidth}
\centering
\includegraphics[width=1.6in]{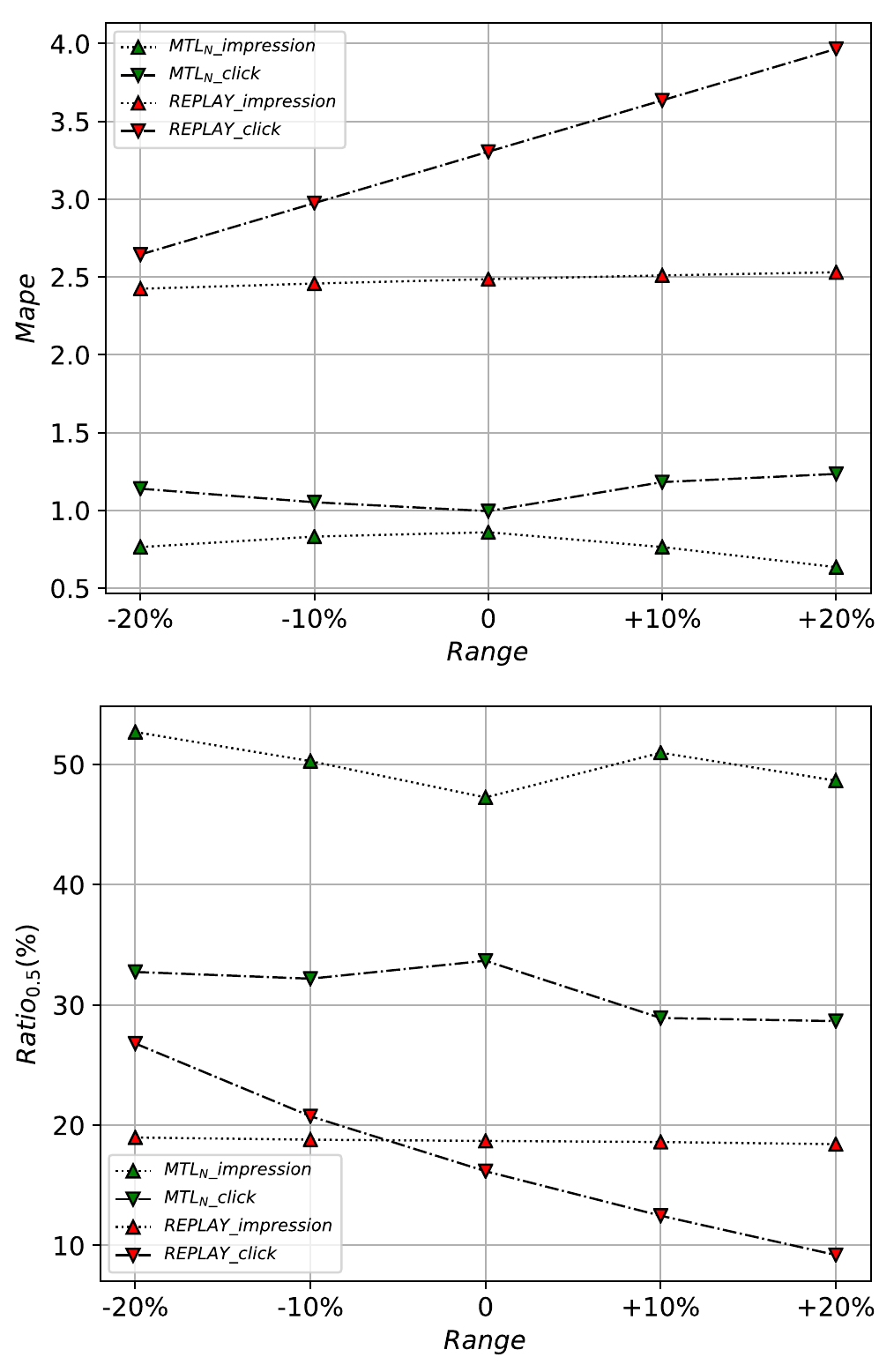}
\end{minipage}%
}%
\centering
\caption{Performance under various $\bm{pctr}$ disturbance.}
\label{pctr_disturbance}
\end{figure}

\vspace{-5pt}
\subsection{Robustness Under Data Disturbance}
As we discussed, due to the dynamic bidding environment and complex strategies, the estimation performance we derive on historical auctions may be unreliable. In this part, we aim to analyze the robustness of the proposed approaches. A natural way to simulate volatility is to add a $pctr$ disturbance in auction logs. Therefore, we adjust the $pctr$ disturbance range($-20\%$$\sim$$+20\%$) on every impression opportunity, and observe the forecasting performance difference between $REPLAY$ and $MTL_N$. On the one hand, we may know if the performance of $REPLAY$ could be good enough under heuristic rules, like $pctr$ adjustment, to discard calibration. On the other hand, calibration robustness of $MTL_N$ should be explored under volatile environments to demonstrate its effectiveness in the real world. As shown in Figure \ref{pctr_disturbance}, $pctr$ disturbances are added to manual bidding and automatic bidding respectively. We have the following observations based on the experiment results.

\begin{itemize}[leftmargin=*]
    \item \textbf{Manual bidding.} For CPC bidding, the bigger $pctr$ is, the higher probability to win an impression. Therefore, the $REPLAY$ method shows huge volatility on all indicators in the experiment, and the overall $mape$ and $ratio_{0.5}$ shows inconsistent changes, thus the optimal heuristic rules are hard to determine. In Contrary, the performance of $MTL_N$ is still surprisingly good whatever the $pctr$ disturbance range, demonstrating the reliability and robustness.
    \item \textbf{Automatic bidding.} No improvement shows on $\#impression$ for automatic bidding, and performance on $\#click$ is proportional to $pctr$ disturbance. We infer that's mainly because the ranking order $\frac{c_i}{d*v_i}$ in retrieved auction set is not changed with the fixed disturbance factor $d$. Similar to manual bidding, the performance of $MTL_N$ is  considerably good in automatic bidding. 
\end{itemize}

Overall, the $REPLAY$ method is sensitive to volatility, and the instability of the method brings great difficulties to the determination of heuristic rules for improving forecasting performance. Moreover, there is far more than one disturbance in the real world, and the disturbance range are different at impression-level either. Compared to $REPLAY$, $MTL_N$ achieves a more robust and much better performance on all indicators across the range of $pctr$ disturbance, demonstrating an excellent calibration robustness to adapt volatile environments.

\subsection{Online Service and Evaluation}
We deploy the framework on Taobao.com for online service. The unified replay algorithm receives campaign criteria from user(means advertiser here) interface and calculates on Maxcompute Hologres \cite{yang2016high}, then the campaign criteria and replay outputs are combined as an HTTP request to call the calibration model on Real-Time Prediction(RTP) center. Finally, calibrated performance is fed to advertisers for campaign optimization. From the latency monitor, most of the forecast requests are responsed in 2 seconds. We conduct a long term A/B test among thousands of advertisers to verify the business value of Campaign Performance Forecasting in the real world.

\begin{itemize}[leftmargin=*]
    \item \textbf{Setups.} We design the following  A/B testing experiment to form fair comparisons. Firstly, we choose long-term active advertisers between 05/29/2020 and 06/30/2020 as experiment candidates, and filter out the outliers with extreme high or low $arpu$ value. Secondly, we sort the candidates with their $arpu$ value, and pick the top $10000$ candidates. Thirdly, we split the $10000$ candidates into A/B two groups with the "ABBA..." order, each group contains $5000$ advertisers. Finally, we analyze the $arpu$ value of two groups between 05/29/2020 and 06/30/2020 to confirm that group A and B have similar $arpu$ performance in A/A test period.  
    
    \item \textbf{Results.} An A/B test starts from 07/01/2020 and runs for 4 weeks. We provide group B with Campaign Performance Forecasting service, and group A stays as is. A $\bm{8}\%$ $arpu$ lift is observed from the online A/B test results after excluding the outliers, demonstrating a significant growth on revenue generation brought by Campaign Performance Forecasting.
     
\end{itemize}

\section{Conclusion and Future Work}
In this paper, we study Campaign Performance Forecasting, which forecasts key performance indicators for a new advertising campaign under given criteria, enabling advertisers manage and optimize their campaigns. The task faces two main technical challenges. (1) How to estimate campaign performance under various bidding types unifiedly, and (2) How to mitigate estimation deviation to provide more accurate forecast results. No prior work studies these challenges yet. To achieve above goals, we propose a novel framework which firstly estimates campaign performance under various bidding types by a unified replay algorithm, and then innovatively introduces a multi-task learning method to calibrate the replay deviation, meeting both interpretability and accuracy requirements. Comprehensive empirical studies demonstrate that our framework significantly outperforms other baselines on a dataset from Taobao.com, and an online A/B test verifies its effectiveness in the real world. To the best of our knowledge, this is the first systematic work for Campaign Performance Forecasting. There are two interesting direction for future study. One is to forecast time series performance for a campaign. The other is to apply campaign performance forecasting to other fields like budget allocation.

\bibliographystyle{ACM-Reference-Format}
\bibliography{sample-sigconf}

\newpage
\appendix

\section{Implementation Details}
In order to improve the reproducibility of our framework, we present essential details for the construction of base log, unified replay algorithm, calibration model, and online latency observation.

\subsection{Log Construction}
\begin{figure}[!htp]
\centering 
\includegraphics[width=0.480\textwidth]{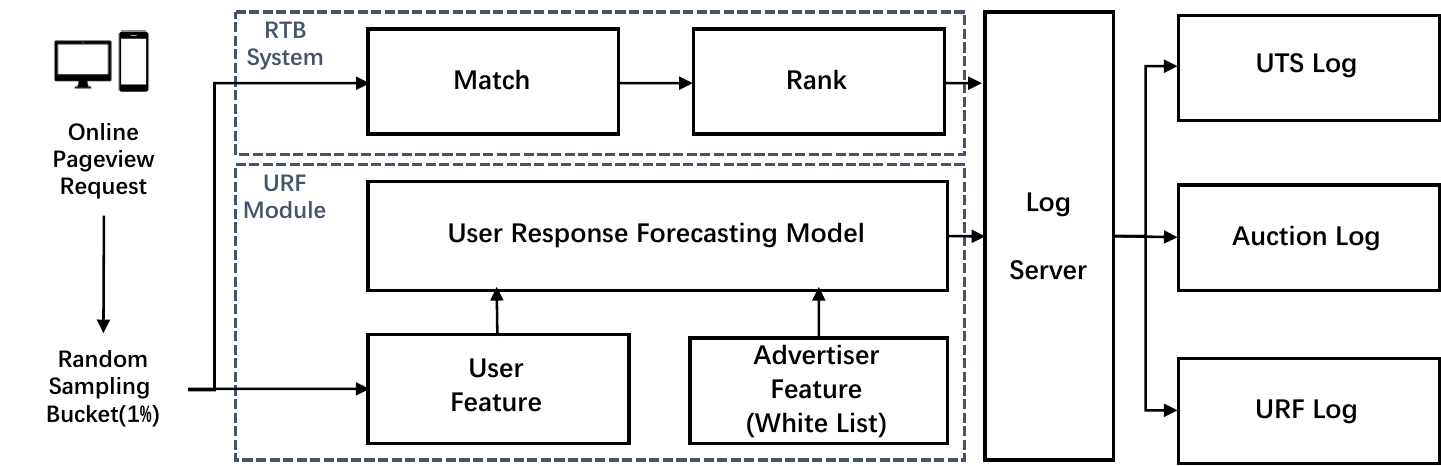}
\caption{The overall log stream.}
\label{Log Stream} 
\end{figure}

\begin{figure}[H]
\centering 
\includegraphics[width=0.480\textwidth]{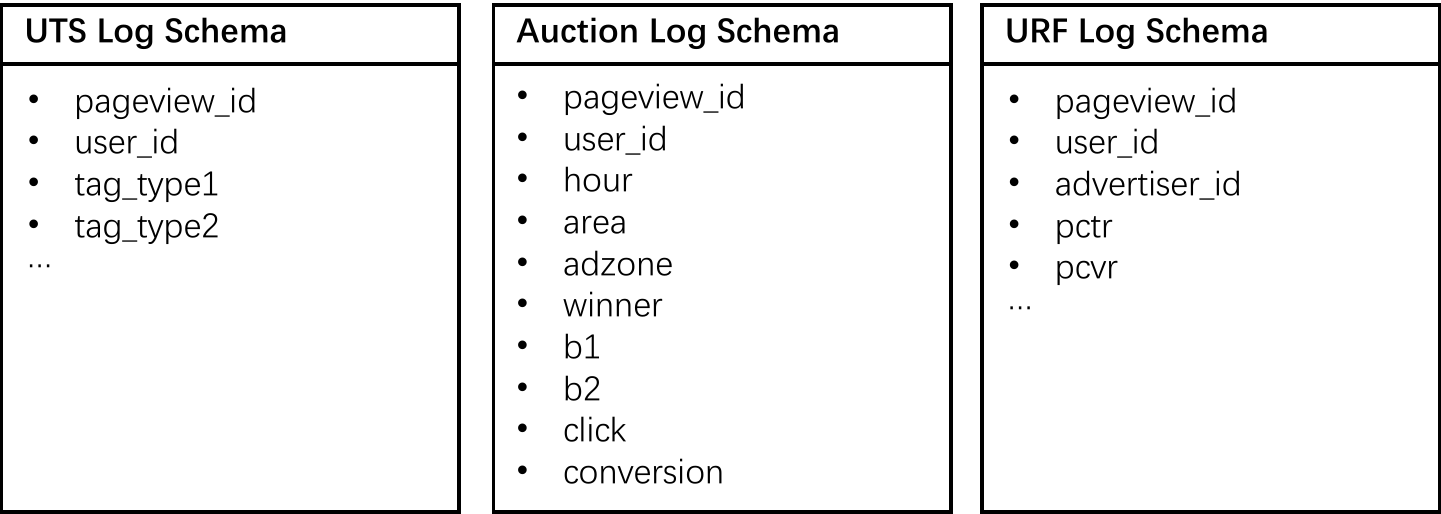}
\caption{Detailed log schema.}
\label{Log Schema} 
\end{figure}

In this section, we introduce how we organize and collect base logs for campaign performance forecasting, and the detailed schema of these logs. For better display of the log construction process, the overall log stream is presented in Figure \ref{Log Stream}, and the detailed log schema is described in Figure \ref{Log Schema}.

\begin{itemize}[leftmargin=*]
    \item \textbf{Down Sampling.} We build a $\frac{1}{100}$ down sampling bucket with a special marker on the online traffic platform, online page view requests arrive continuously while users log in and browse on Taobao.com. The probability of entering the bucket is equal for every request, those who pass through the bucket will be always marked in the following log stream. It should be noted that the bucket marker on requests plays an important role in computation complexity and latency optimization in our framework.
    \item \textbf{UTS Log and Auction Log.} All the ad requests will go through the real-time bidding(RTB) system for auction, in which match module retrieves all targeted bidders, and rank module calculates the bid price for all bidders and rank to decide the final winner.  In match phase, the targeting types and values from bidders which retrieves a request will be recorded into User-Tag-Service(UTS) Log. The context, winner, rank score and corresponding clicks/conversions of a request will be recorded as Auction Log either. In final deployment, the UTS and Auction Logs without the bucket marker are filtered out to optimize computation complexity.
    \item \textbf{URF Log.} When ad requests arrives at User Response Forecasting(URF) module, the module will retrieve essential features for marked requests and white-list advertisers, and then feed into trained URF model to predict $pctr$ and $pcvr$ values. All the outputs will be written into URF Log. Notably, the deployed URF model here is trained on Action Log, and the white list on advertisers is designed for online A/B test.
\end{itemize}

As most advertising platforms have similar bidding system to Taobao.com, the log construction in our framework could be well generalized to others.

\begin{figure}[H]
\centering 
\includegraphics[width=0.480\textwidth]{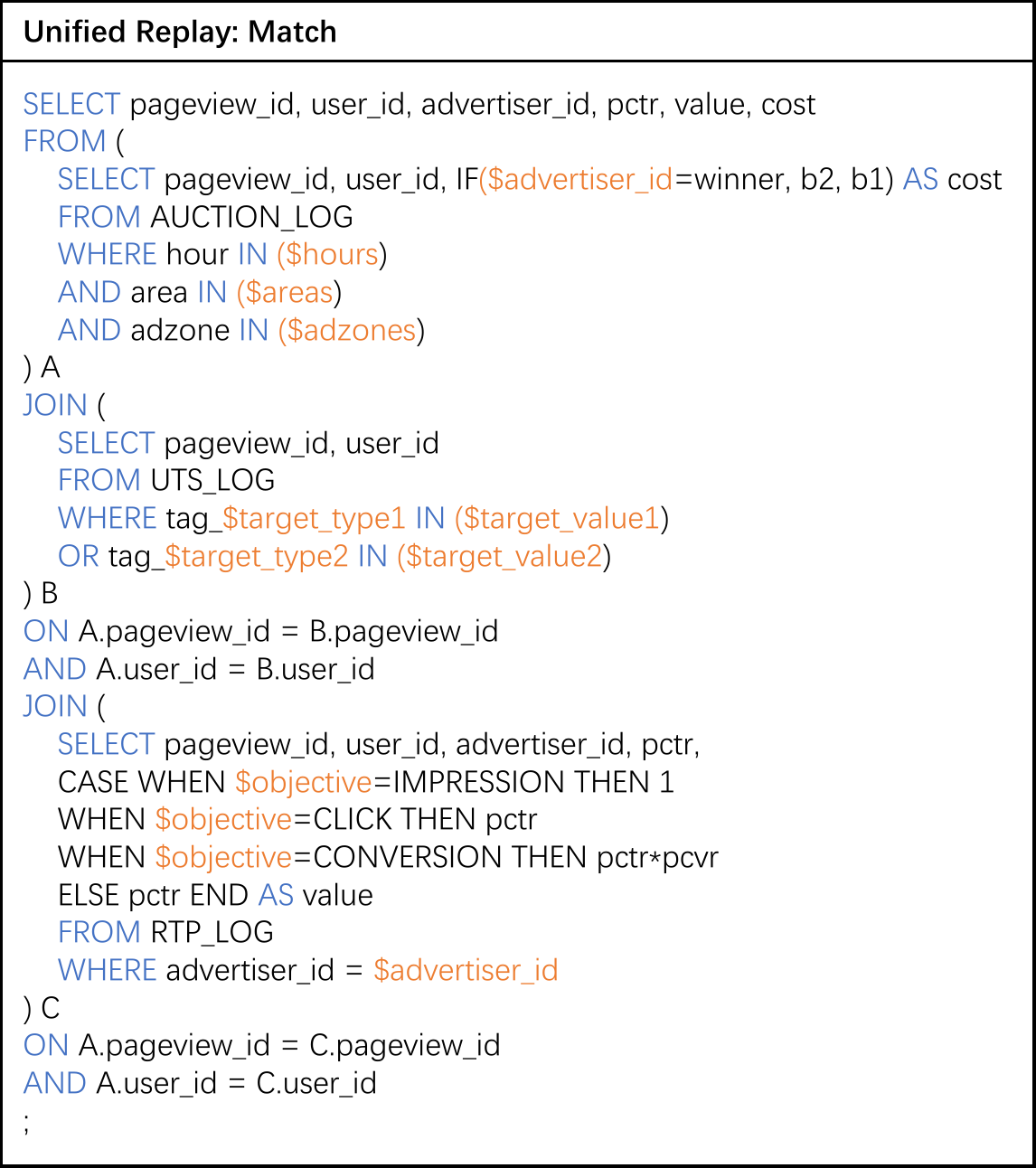}
\caption{The SQL logic of Match Phase, variables in orange are inputs from given campaign criteria.}
\label{Match SQL} 
\end{figure}

\subsection{Unified Replay Algorithm}
In this section, we show more additional details about the unified replay algorithm that are essential in supporting our reproducibility. With the base logs as described in Appendix A, we focus on the elaboration of 'Match Phase' and 'Rank Phase' in replay and the latency optimization for online deployment.

\begin{figure}[!htb]
\centering 
\includegraphics[width=0.480\textwidth]{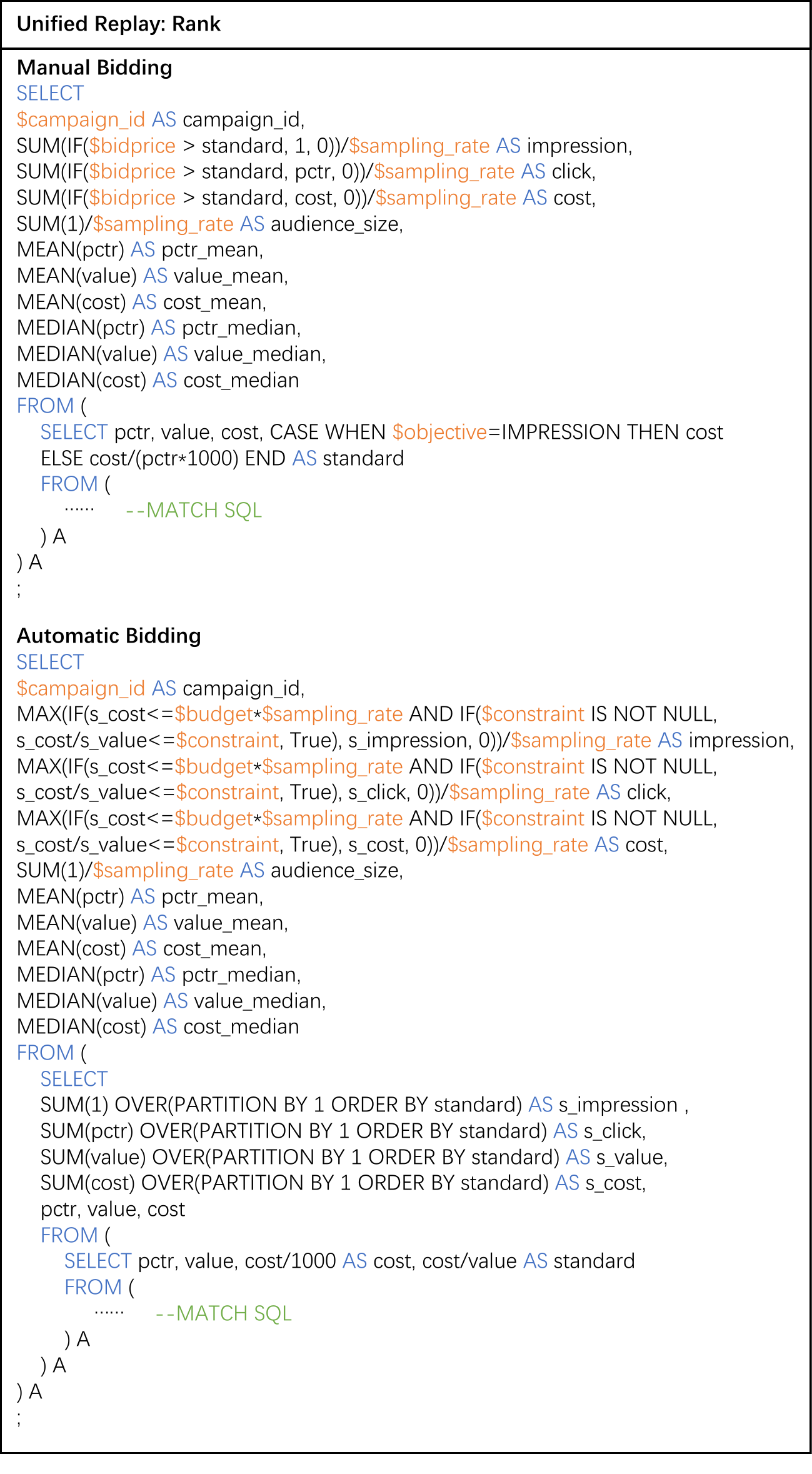}
\caption{The SQL logic of Rank Phase, including both manual bidding and automatic bidding. Variables in orange are inputs from given campaign criteria.}
\label{Rank SQL} 
\end{figure}

\begin{figure}[!htb]
\centering 
\includegraphics[width=0.495\textwidth]{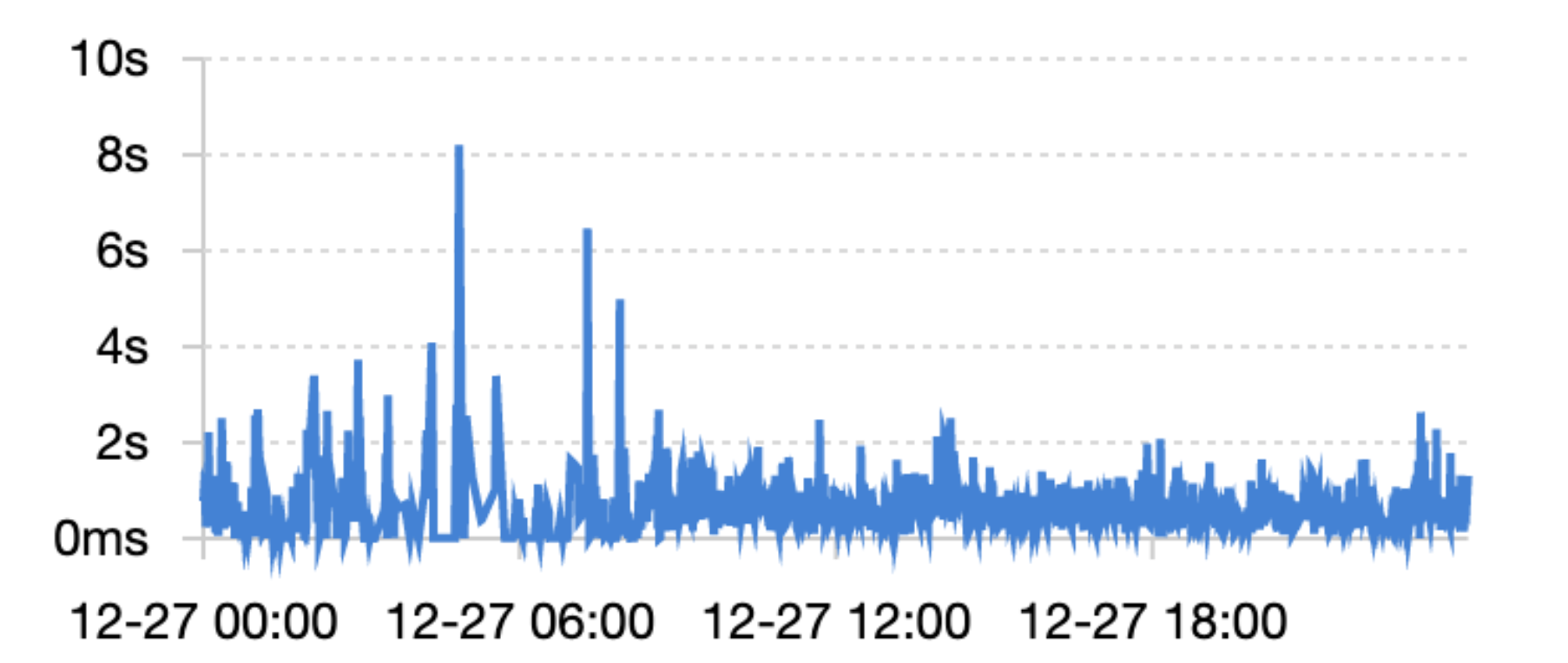}
\caption{The latency of online service.}
\label{Latency} 
\end{figure}

\begin{itemize}[leftmargin=*]
    \item \textbf{Match Phase.} Match phase aims to retrieve complete auction information from historical logs after advertisers specify their campaign criteria. The detailed logic is shown in Figure \ref{Match SQL}.
    \item \textbf{Rank Phase.} Based on the auction information retrieved in Match Phase, Rank phase determines the final auctions in which the campaign would win, and then calculate the performance for campaigns. The detailed logic is shown in Figure \ref{Rank SQL}.  
    \item \textbf{Online Latency.} In the proposed framework, most of the computation is concentrated in the replay algorithm, we adopt several speed-up actions to optimize the latency for online service. (1) As described in Appendix A, a down sampling rate $\frac{1}{100}$ is applied in log stream to reduce computation complexity. (2) Only essential Match Phase and Rank Phase in auction process are considered in replay, those strategies which are hard to reproduce are neglected, leaving the replay deviation to be calibrated in performance calibration module. (3) Engineering optimization on MaxCompute Hologres\cite{yang2016high}. We present the online service latency monitor of 27/12/2020 in Figure \ref{Latency}, it's clear that most requests are responded in 2 seconds.
\end{itemize}

\begin{table}[!htb]
\caption{Input of Calibration Model.}
\label{table:calibration input construction}
\centering
\footnotesize 
\resizebox{\linewidth}{!}{%
\begin{tabular}{lp{6.0cm}}
\hline 
\textbf{Domain} & \textbf{Element} \\
\hline
\tabincell{c}{Campaign criteria} & \tabincell{l}{ targeting\_option, objective, bidding\_type.} \\
\hline
\tabincell{c}{Replay outputs} & \tabincell{l}{pctr\_mean, pctr\_median, cost\_mean, cost\_median, cost, \\value\_mean, value\_median, audience\_size, click, impression.}\\ 
\hline
\end{tabular}
}
\end{table}

\subsection{Calibration Model}
In this section, we mainly elaborate the construction of calibration model. We first present our consideration on feature selection, then the implementation details of offline training and online deployment respectively.

\begin{itemize}[leftmargin=*]
    \item \textbf{Feature Selection.} The calibration features consist of two parts, campaign criteria and replay outputs. For campaign criteria, targeting option, objective and bidding type are usually tied to delivery strategies, thus we choose these criteria to capture calibration patterns. For replay outputs, statistical features in match phase are calculated to represent the targeting crowd quality for a campaign, and the final results in rank phase are applied as base campaign performance. As detailed campaign criteria is described in Section 3.1, and replay outputs of Match Phase and Rank Phase are clearly illustrated in Algorithm 1, we list the overall input for calibration model in Table \ref{table:calibration input construction}.
    \item \textbf{Offline Training.} The replay algorithm are accomplished on the high-speed distributed cloud computing frame MaxCompute Hologres\cite{yang2016high}. We collect campaigns samples as described in Section 4.1. Campaign criteria, replay outputs and  the true performance are adopted as features and labels respectively for offline training, and the discrete values in campaign criteria are processed to 1-hot vectors. It should be noticed that most settings for model training follow the description in \cite{guo2017deepfm,ma2018modeling}. The number of tasks and experts are set to $3$ and $6$, hidden units of expert layer and tower layer are set to $64$, $32$ respectively. Batch-Norm layers\cite{ioffe2015batch} are adopted in the models, and Adam\cite{kingma2014adam} is adopted for optimization with a initial learning rate=$10^{- 3}$. Our code is implemented with TensorFlow\cite{abadi2016tensorflow} in python. 
    \item \textbf{Online Service.} The trained model is deployed on the Real-Time Prediction(RTP) center in Taobao advertising system for online service. The unified replay algorithm receives campaign criteria from user interface and calculate the results in real-time on MaxCompute\cite{yang2016high}, then the campaign criteria and replay outputs are combined as input to call the calibration model by an HTTP request. Finally, calibrated performance is fed to advertisers for campaign optimization.
\end{itemize}

\end{document}